\documentclass{article}

\usepackage{microtype}
\usepackage{graphicx}
\usepackage{subfigure}
\usepackage{booktabs} 
\usepackage{amsmath}
\usepackage{amssymb}
\usepackage{caption}
\usepackage{array}

\usepackage{hyperref}

\usepackage[accepted]{icml2020}

\icmltitlerunning{Unbalanced GANs: Pre-training the Generator of Generative Adversarial Network using Variational Autoencoder}

\begin{document}

\twocolumn[
\icmltitle{Unbalanced GANs: Pre-training the Generator of Generative \\ 
Adversarial Network using Variational Autoencoder}



\icmlsetsymbol{equal}{*}

\begin{icmlauthorlist}
\icmlauthor{Hyungrok Ham}{kaist}
\icmlauthor{Tae Joon Jun}{asan}
\icmlauthor{Daeyoung Kim}{kaist}
\end{icmlauthorlist}

\icmlaffiliation{kaist}{School of Computing, KAIST, Daejeon, South Korea}
\icmlaffiliation{asan}{Asan Medical Center, Seoul, South Korea}
    
\icmlcorrespondingauthor{Hyungrok Ham}{gudfhr95@kaist.ac.kr}

\icmlkeywords{Machine Learning, Deep Learning, Generative Adversarial Network, Variational Autoencoder, ICML}

\vskip 0.3in
]



\printAffiliationsAndNotice{}  

\begin{abstract}
We propose Unbalanced GANs, which pre-trains the generator of the generative adversarial network (GAN) using variational autoencoder (VAE). We guarantee the stable training of the generator by preventing the faster convergence of the discriminator at early epochs. Furthermore, we balance between the generator and the discriminator at early epochs and thus maintain the stabilized training of GANs. We apply Unbalanced GANs to well known public datasets and find that Unbalanced GANs reduce mode collapses. We also show that Unbalanced GANs outperform ordinary GANs in terms of stabilized learning, faster convergence and better image quality at early epochs. 
\end{abstract}

\section{Introduction}
\label{introduction}
Generative models have come a long way. Researchers proposed various generative models such as Restricted Boltzmann Machine (RBM) \cite{hinton2006reducing}, Deep Boltzmann Machine (DBM)  \cite{salakhutdinov2009deep}. Among these generative models, Variational Autoencoder (VAE) \cite{kingma2013auto} and Generative Adversarial Network (GAN) \cite{goodfellow2014generative} are the most popular models. \\
Recently, GAN is under the research spotlight since it produces sharp images and better image quality than any other generative models. Accordingly, researchers proposed variants of GANs such as architectural variants e.g. Deep Convolutional GAN (DCGAN) \cite{radford2015unsupervised}, Energy-Based GAN (EBGAN) \cite{zhao2016energy} or loss variants e.g. Least Squares GAN (LSGAN) \cite{mao2017least}, Wasserstein GAN (WGAN) \cite{arjovsky2017wasserstein}. \\
\begin{figure}[ht]
\vskip 0.2in
\begin{center}
\centerline{\includegraphics[width=\columnwidth]{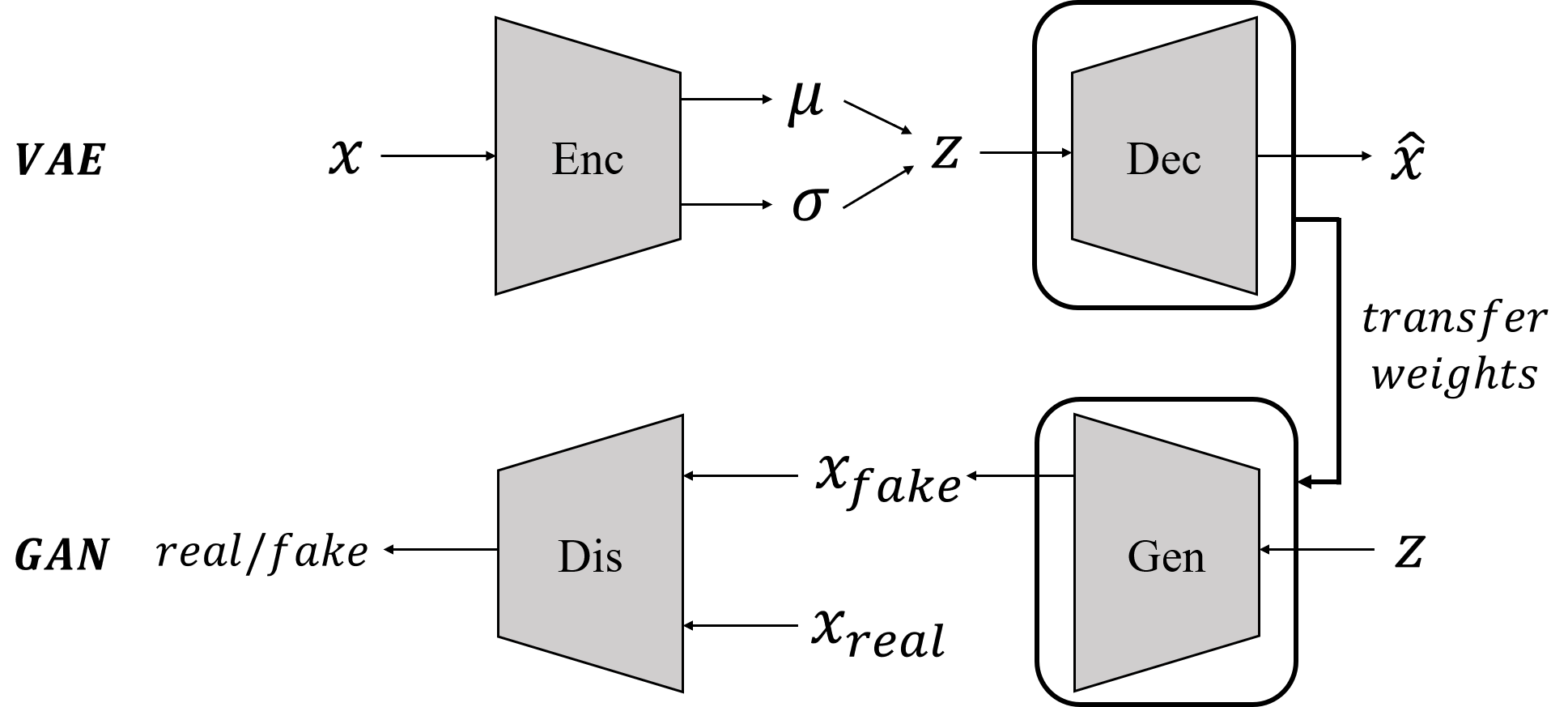}}
\caption{Overview of our method. We first train VAE and transfer the weights of the variational decoder to the generator. Then, train GAN with the pre-trained generator.}
\label{fig:figure1}
\end{center}
\vskip -0.2in
\end{figure} 
However, GAN and its variants have innate disadvantages which make GANs hard to train. It is difficult to balance the learning speed of the generator and the discriminator so the losses of these two networks oscillate. Besides, GANs often suffer from mode collapse. Among these disadvantages, what we want to emphasize is balancing the generator and the discriminator. Balancing the generator and the discriminator is hard because in many cases the discriminator converges faster than the generator. \\
To solve this phenomenon, Boundary Equilibrium GAN (BEGAN) \cite{berthelot2017began} proposed equilibrium hyperparameter that defines the ratio between losses of the generator and the discriminator. BEGAN showed impressive performance with a simple architecture. BEGAN proved that balancing the generator and the discriminator is important for GANs performance. However, BEGAN architecture is not widely applicable since it has a different structure from other GANs. BEGAN used EBGAN based architecture which uses autoencoder as a discriminator and consequently uses pixel-wise mean squared loss. \\
In this paper, we propose Unbalanced GANs, which uses pre-trained variational decoder as a generator of GAN. There are similar attempts with us which is to combine VAE and GAN. One of the examples is VAEGAN \cite{larsen2015autoencoding}. VAEGAN uses an end-to-end architecture that is configured in the order of variational encoder, decoder, and discriminator. This architecture makes possible to generate sharper and more realistic images than VAE. However, since VAEGAN trains three networks simultaneously, it is difficult to extract the latent distribution in the variational encoder and discriminate between the real and generated images in the discriminator. \\
During the training process, first, we train VAE with a given dataset. Then, we transfer the weight of the variational decoder to the generator. With the pre-trained generator, we train GAN again in the same way as before. Our method can be used in any GANs. If there is a pre-trained variational decoder which has the same network architecture with a generator, then our method can be utilized and enhance GANs' performance. \\
Since VAE is trained while assuming the prior distribution as the normal distribution, the approximated posterior distribution also becomes the normal distribution. If we initialize the weights of the generator using the pre-trained variational decoder, we can start training GAN in the state that the generative distribution is equal to the normal variational decoder distribution. due to this, we can prevent the fast convergence of the discriminator and stabilize the training of GAN. \\
Our contributions are as follows: 
\begin{itemize}
\item We combine VAE with GAN by using a pre-trained variational decoder as a generator of GAN. Our method can be applied to any GANs by constructing proper VAE architecture according to GAN's.
\item Using pre-trained generator, we prevent the discriminator from winning too easily (faster convergence of the discriminator compare to the generator) at the early epoch while maintaining the stabilized learning process of GANs and reducing mode collapses. 
\item We show faster convergence of the generator and the discriminator and better image quality at early epochs compared to ordinary GANs.
\end{itemize}

\section{Related Work}
\label{relatedwork}
\subsection{Variational Autoencoder}
VAE has two networks, an encoder and a decoder network. The encoder network ${q}$ samples the distribution ${z}$ of a given data ${x}$ assuming that the distribution of the dataset is normal ${N(0, I)}$. The decoder network ${p}$ reconstructs the data ${x}$ from the sampled distribution ${z}$. 
\begin{equation}
    \begin{split}
        & z \sim Enc(x) = q(z \mid x) \\
        & \hat{x} \sim Dec(z) = p(x \mid z)
    \end{split}
\end{equation}
VAE uses variational inference to approximate prior distribution. The VAE loss is a mixture of the prior regularization parameter and the reconstruction loss:
\begin{equation}
    \begin{split}
        & \mathcal{L}_{VAE} =  \mathcal{L}_{reg} + \mathcal{L}_{recon} \\
        & \mathcal{L}_{reg} = D_{KL}(q(z \mid x)) \mid\mid p(z)) \\
        & \mathcal{L}_{recon} = -\mathbb{E}_{q(z \mid x)}[\log p(x \mid z)] \\
    \end{split}
\end{equation}
where ${D_{KL}}$ is the Kullback–Leibler (KL) divergence. \\ 
The reconstruction loss is pixel-wise Binary Cross Entropy (BCE) between real images $x$ and reconstructed images $\hat{x}$. The regularization parameter regularizes the distribution of $q$ to be the zero-mean normal distribution by minimizing KL divergence. \\
VAE is stable during training because VAE uses pixel-wise reconstruction error. However, as VAE is optimized to match the average reconstruction loss of given inputs, it produces blurry images. \\ 
If we use a pre-trained variational decoder as a generator, we can utilize the advantages of both VAE and GAN. We can initialize the generative distribution with the variational decoder distribution which is the normal distribution. Through training GAN, we can also make blurry images of VAE sharp and clear. Furthermore, there is no concern about the failure of pre-training.

\subsection{Generative Adversarial Network}
GAN has two networks: a generator and a discriminator network. GAN is trained by a two-player game between the generator and the discriminator. The generator network ${G}$ creates samples to fool the discriminator. The discriminator network ${D}$ examines samples to determine whether the given input is real or generated. \\
Given random noise ${z \sim p_z(z)}$ and real data ${x}$, the generator is trained to minimize ${\log  (1-D(G(z))}$ and the discriminator is trained to maximize ${\log D(x) \in [0, 1]}$
\begin{equation}
    \begin{split}
        &\mathcal{L}_{GAN} = \min\limits_{G} \max\limits_{D} V(G, D) \\
        & = \mathbb{E}_{x \sim p_{{data}(x)}}[\log(D(x))] + \mathbb{E}_{z \sim p_{z}(z)}[\log(1-D(G(z)))]
    \end{split}
\end{equation}
where ${x \sim p_{{data}(x)}}$ and ${z \sim p_{z}(z)}$ is the real data distribution and the prior distribution respectively. GAN usually uses BCE loss for the generator and the discriminator. However, some variants of GANs use different losses such as mean squared error. \\
When the generator is fixed, the optimal discriminator is
\begin{equation}
    D^{*}(x) = {\frac{p_{data}(x)}{p_{data}(x)+p_{g}(x)}}
\end{equation}
And given optimal discriminator, GAN loss has its global optimum when ${p_{g} = p_{data}}$. At that point, the loss of the generator becomes
\begin{equation}
    C(G) = -\log(4) + 2 \times JSD(p_{data} \mid\mid p_{g}) 
\end{equation}
where JSD is the Jensen-Shannon (JS) divergence. \\ 
This means that the generator is trained to minimize the distance between prior and real data distribution. As a result, the generator learns to mimic real data distribution. \\
GAN is one of the most promising generative models since it produces sharp and diverse images. However, because GAN loss is a min-max loss and GAN trains the generator and the discriminator alternately, it is hard to reach global optimum which is a saddle point. To successfully reach the saddle point, balancing the generator and the discriminator is essential. But the discriminator usually converges faster than the generator so it is hard to attain the balance between them. If GAN fails to balance between them, GAN might fail to learn or suffer from mode collapse.

\subsection{Boundary Equilibrium GAN}
BEGAN proposed an equilibrium enforcing method between the loss of the generator and the discriminator. BEGAN used autoencoder as a discriminator which was first proposed in EBGAN. While other GANs attempted to match data distribution directly, BEGAN matched the autoencoder loss distributions between real and generated data by minimizing the Wasserstein distance. \\
To maintain the balance between the generator and the discriminator loss, BEGAN introduced an equilibrium hyperparameter ${\gamma \in [0,1]}$:
\begin{equation}
    \gamma = {\frac{\mathbb{E}[\mathcal{L}(G(z))]}{\mathbb{E}[\mathcal{L}(x))]}}
\end{equation}
Besides, BEGAN borrowed the idea of Proportional Control Theory and used a variable $k_t$ to maintain the equilibrium $\mathbb{E}[\mathcal{L}(G(z))] = \gamma\mathbb{E}[\mathcal{L}(x))]$. BEGAN loss is as follows:
\begin{equation}
    \begin{cases}
        \mathcal{L}_{D} = \mathcal{L}(x) - k_{t} \times \mathcal{L}(G(z_{D}))\\
        \mathcal{L}_{G} = \mathcal{L}(G(z_{G}))\\
        k_{t+1} = k_{t} + \lambda_{k}(\gamma\mathcal{L}(x) - \mathcal{L}(G(z_{G})))
    \end{cases}
\end{equation}
where $\mathcal{L}$ is a pixel-wise autoencoder loss
\begin{equation}
    \mathcal{L}(v) = {\mid v - D(v) \mid}^\eta \qquad \eta \in \{1, 2\}
\end{equation}
The discriminator of BEGAN has two objectives: autoencoding and reconstructing real images, and discriminating real images from generated images. BEGAN can balance between these two objectives by adjusting ${\gamma}$ and maintain the ratio of the generator and the discriminator loss. \\
By introducing ${\gamma}$, BEGAN showed remarkable performance yet with simple architecture. BEGAN pointed out that balancing the generator and the discriminator is crucial for improving GANs performance. \\
However, BEGAN loss and $\gamma$ may not be applied to other GANs since BEGAN uses pixel-wise autoencoder loss. In order to use them, it is forced to use autoencoder as a discriminator and various structures of the discriminator may not be used. Additionally, as BEGAN is trained to match the distribution of autoencoder losses of ground truth data, it cannot solve mode collapse.

\subsection{Transferring GANs}
Transferring knowledge of pre-trained network and fine-tuning are widely applied techniques to enhance discriminative models' performance. Adopting these, Wang et al. \yrcite{wang2018transferring} studied about transferring pre-trained knowledge on GANs and proposed Transferring GANs.\\
They experimented transferring knowledge on WGAN-GP \cite{gulrajani2017improved}. WGAN-GP has ResNet \cite{he2016deep} based architecture and is known to be stable and robust. Batch normalization \cite{ioffe2015batch} and layer normalization \cite{ba2016layer} are also used in both generator and discriminator. \\ 
They divided datasets into source and target domains. The source domain is the dataset that they pre-trained the network on and the target domain is the dataset that pre-trained GANs were adapted on. They first pre-trained four WGAN-GP networks on four source datasets. Then, they randomly chose the target dataset and automatically estimate the most suitable pre-trained network by calculating Fr\'etchet Inception Distance (FID) \cite{heusel2017gans} between the source and the target dataset. Next, they transferred the knowledge of the most suitable network to the chosen dataset. \\
They experimented every combination which is initializing the generator and the discriminator with random or pre-trained weights and concluded that initializing both networks with pre-trained weights obtained the best result. \\
Transferring GANs showed that transferring knowledge and domain adaptation can also be used in not only discriminative models but also generative models. Furthermore, they improved the performance of GANs in terms of faster convergence and improving the quality of generated images. \\
However, since transferring GANs used GAN to pre-train GAN, it is not free with the innate difficulties of GANs. If they fail to pre-train GANs on source datasets while suffering from oscillating or mode collapse, then it would be difficult to transfer knowledge or adapt the pre-trained network to the target domain.

\section{Method}
\label{method}
In this section, we provide our methods to train our model Unbalanced GANs. We explain why we choose the variational decoder to pre-train the generator. We also provide details of training process and model architecture.

\subsection{Variational Decoder as a Generator}
There are mainly four reasons why we use the variational decoder to pre-train the generator. \\
First, VAE and GAN use similar losses to find prior distribution. VAE loss consists of two losses: regularization and reconstruction loss. Reconstruction loss is pixel-wise BCE between real images and reconstructed images. Regularization loss is the KL divergence between a prior distribution and a sampled distribution. While minimizing the VAE loss, regularization loss is also minimized and thus the KL divergence between two distributions decreases. GAN loss is a min-max loss between the generator and the discriminator. When the discriminator reaches its global optimum, GAN loss becomes the JS divergence between real and generated data distribution. As the JS divergence and the KL divergence are metrics to examine the distance between two distributions, VAE and GAN have a point of sameness that they are trained to minimize the distance between two distributions. Besides, minimizing the KL divergence affects minimizing the JS divergence because the JS divergence is the symmetric version of the KL divergence. \\
Second, the generator distribution can be initialized with the normal distribution. Since VAE assumes that the prior distribution is the normal distribution, the approximation class of the posterior distribution is trained to match the normal distribution and becomes similar to it. On the other hands, GAN does not assume any distribution from the beginning of the training so the generator distribution is trained to approximate any distribution. By initializing the generator distribution with the normal distribution, we can prevent the generator from converging to a strange distribution and suffering from mode collapse. Additionally, we can prevent the discriminator from winning too easily at early epochs. As the generator distribution is initialized with the variational decoder distribution, the generator generates blurry images at the beginning. However, after several epochs, the generator distribution is approximated to match the ground truth data distribution and it produces sharp images due to the GAN loss. \\
Third, VAE is stable. Unlike GAN, as VAE uses pixel-wise BCE loss (reconstruction loss) between real and reconstructed images, the variational encoder and the decoder network is trained to converge to the local minimum. There is no need to balance these networks. Hence, VAE does not suffer from mode collapse or loss oscillating. If we use GAN to pre-train GAN, there is a chance to fail pre-training. However, the use of VAE for pre-training can fundamentally prevent pre-training from failure. \\
Fourth, VAE and GAN both can use the normal distribution to sample noises which are used as input of the variational decoder and the generator. VAE uses a reparametrization trick to sample latent distribution on the normal distribution. GAN indeed uses the normal distribution to sample latent noises. If they are using different types of inputs, it will be difficult to transfer weights. However, since they are having the same type of inputs, we can transfer the weights of the variational decoder to the generator.

\begin{algorithm}[tb]
    \caption{Training Unbalanced GANs}
    \label{alg:unbalancedgans}
    \begin{algorithmic}
        \STATE // Train VAE
        \STATE ${\theta_{Enc}, \theta_{Dec}}$~//~initialize~weights~of~VAE
        \FOR{i steps}
            \STATE ${X \leftarrow random~minibatch~from~dataset}$
            \STATE ${Z_{VAE} \leftarrow Enc(X)}$
            \STATE ${\mathcal{L}_{reg} \leftarrow D_{KL}(q(Z_{VAE} \mid X)) \mid\mid p(Z_{VAE}))}$
            \STATE ${\hat{X} \leftarrow Dec(Z_{VAE})}$
            \STATE ${\mathcal{L}_{recon} \leftarrow -\mathbb{E}_{q(Z_{VAE} \mid X)}[\log p(X \mid Z_{VAE})]}$ 
            \STATE ${\theta_{Enc} \xleftarrow{+}  - \nabla_{\theta_{Enc}}(\mathcal{L}_{reg} + \mathcal{L}_{recon})}$
            \STATE ${\theta_{Dec} \xleftarrow{+}  - \nabla_{\theta_{Dec}}(\mathcal{L}_{reg} + \mathcal{L}_{recon})}$
        \ENDFOR
        \STATE // Train GAN
        \STATE ${\theta_{Dis}}$~//~initialize~weights~of~discrimiantor
        \STATE ${\theta_{Gen} \leftarrow \theta_{Dec}}$~//~transfer weights of decoder to generator
        \FOR{j steps}
            \STATE ${X \leftarrow random~minibatch~from~dataset}$
            \STATE ${Z_{GAN} \leftarrow samples~from~N(0,I)}$
            \STATE ${\mathcal{L}_{Dis} \leftarrow \mathbb{E}[\log(D(X)) + \log(1-D(G(Z_{GAN})))]}$
            \STATE ${\theta_{Dis} \xleftarrow{+} +\nabla_{\theta_{Dis}}(\mathcal{L}_{Dis})}$
            \STATE ${\mathcal{L}_{Gen} \leftarrow \mathbb{E}[\log(1-D(G(Z_{GAN})))]}$
            \STATE ${\theta_{Gen}  \xleftarrow{+} -\nabla_{\theta_{Gen}}(\mathcal{L}_{Gen})}$
        \ENDFOR
    \end{algorithmic}
\end{algorithm}

\subsection{Training Unbalanced GANs}
Training Unbalanced GANs is a sequential composition of training VAE and GAN. In Algorithm~\ref{alg:unbalancedgans}, we provide pseudocode of the training procedure of Unbalanced GANs. \\
We first set a target GAN (DCGAN, LSGAN, WGAN) to train and design architecture of it. Then, based on the target GANs' architecture, we construct an architecture for VAE. We provide the details of architecture in Section \ref{architecture}. \\
Next, we train and update the weights of VAE with a given dataset for several epochs. After training, VAE can generate blurry images similar with the dataset. Using pre-trained weights and knowledge, we transfer the weights and knowledge of the variational decoder to the generator. Finally, we train and update the weight of GAN for several epochs. \\
The transferred weights of the variational decoder enable the generator to produce blurry but similar images with the dataset at the beginning of the training. This prevents the discriminator to overwhelm the generator at the beginning and thus can balance the generator and the discriminator at the early stage of the training.

\begin{figure*}[ht]
\vskip 0.2in
\begin{center}
\centerline{\includegraphics[width=6in]{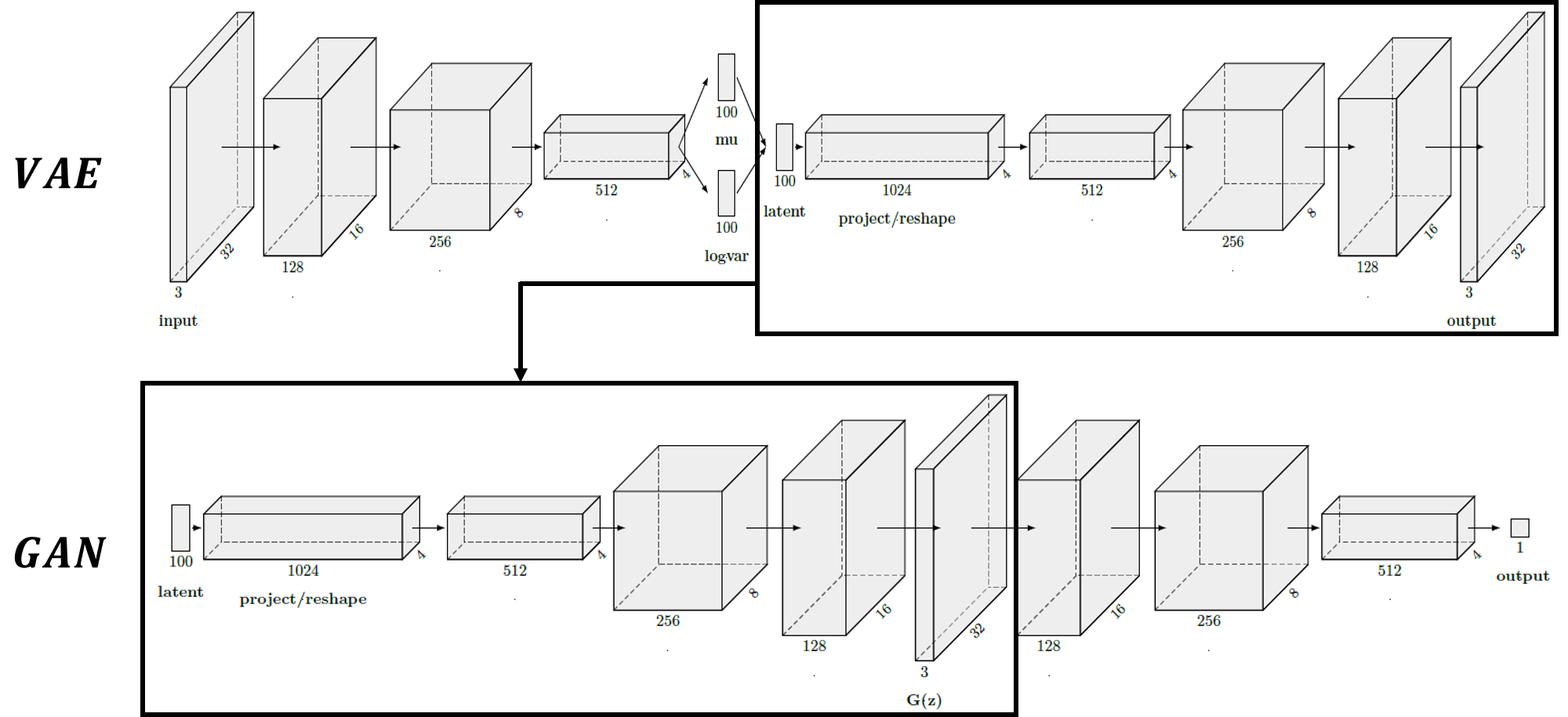}}
\caption{Sample architecture for training the Unbalanced GAN on the CIFAR-10 dataset. Since we transfer the weights of the variational decoder to the generator, they have the same architecture. Furthermore, the variational encoder and the discriminator have the same architecture except for the output layer.}
\label{figure2}
\end{center}
\vskip -0.2in
\end{figure*}

\subsection{Model Architecture}
\label{architecture}
For the model architecture of GAN, we mainly follow the architectural guidelines for Deep Convolutional GANs. 3 $\times$ 3 convolutional layers with stride 2 are used for downsampling and 3 $\times$ 3 fractional-strided convolutional layers with stride 2 are used for upsampling. Leaky ReLU \cite{maas2013rectifier} activation function is used in hidden layers of the variational decoder and the generator network, and ReLU \cite{nair2010rectified} activation function is used in hidden layers of the variational encoder and the discriminator. Besides, we use the Tanh activation function for the output layer of the generator and the Sigmoid activation function for the output layer of the discriminator. We applied dropout layers \cite{srivastava2014dropout} at the variational encoder and the discriminator to prevent overfitting. Additionally, we use batch normalization between all layers except for the output layer of the generator and the input layer of the variational encoder and the discriminator. \\
The architecture of VAE is similar to the target GAN. As we transfer the weights of the variational decoder to the generator, the variational decoder has the same architecture with the generator. The variational encoder also has a similar architecture with the discriminator except for the output layer. The variational encoder has a fully-connected output layer having the same number with the latent dimension to sample the latent distribution. The discriminator has only one fully-connected output layer with the Sigmoid activation function to calculate the probability that the given input is real or generated. We display one of the sample architectures of Unbalanced GANs in Figure~\ref{figure2}. \\
The main idea of Unbalanced GANs is to pre-train the generator using the variational decoder. Our proposed method can be adopted by any GANs. The architecture of Unbalanced GANs can be varied according to the generator architecture that you want to pre-train. When you construct your own Unbalanced GANs, you have to properly construct VAE to pre-train. It is recommended to use the same architecture for variational decoder and generator, variational encoder and discriminator.

\section{Experiments}
\label{experiments}
\subsection{Setup}
We trained three Unbalanced GANs: DCGAN, LSGAN and WGAN on MNIST \cite{lecun1998gradient}, CIFAR-10 \cite{krizhevsky2009learning} and LSUN Bedroom \cite{yu15lsun} dataset. We chose these GANs since to the best of our knowledge almost all variants of GANs are modified versions of them. We followed the main design concepts of each GANs when implementing.  \\
We trained our models using Adam optimizer \cite{kingma2014adam} with learning rate 0.0002, $\beta_1$ 0.5 and $\beta_2$ 0.999. We applied the same hyperparameters at both GAN and VAE. We used 128 for the latent dimension, 64 for the batch size. In the Leaky ReLU activation function, we used 0.2 for the slope of the leak. The dropout rate was set to 0.25. We initialized the weights of GAN and VAE with zero-centered normal distribution with standard deviation 0.02 except for the generator which is initialized with the transferred weights of the variational decoder. \\
We trained our models with 28, 32, 64 image sizes according to the dataset, adding or removing convolutional layers. Training was on two NVIDIA TITAN X GPUs. We pre-trained VAE about 50000 steps and trained GAN about the same number of steps with VAE.

\begin{table*}[t]
    \vskip 0.1in
    \setlength\heavyrulewidth{0.25ex}
    \begin{small}
    \begin{center}
        \begin{tabular}{ccccc}
            \toprule
            & \multicolumn{2}{c}{DCGAN} & \multicolumn{2}{c}{LSGAN} \\
            \midrule
            & Generator & Discriminator & Generator & Discriminator \\
            \toprule
            & \includegraphics[width=1.25in]{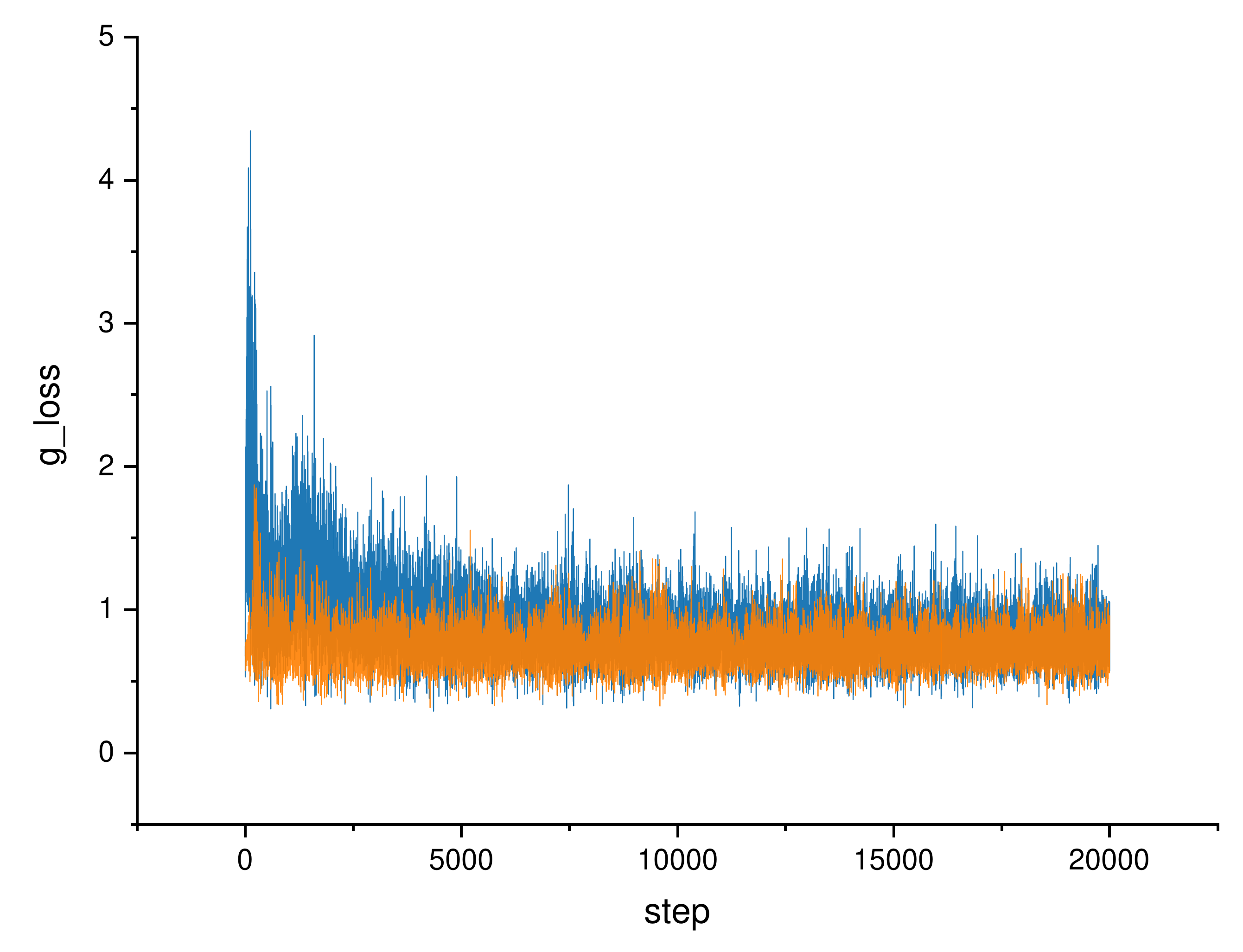} 
            & \includegraphics[width=1.25in]{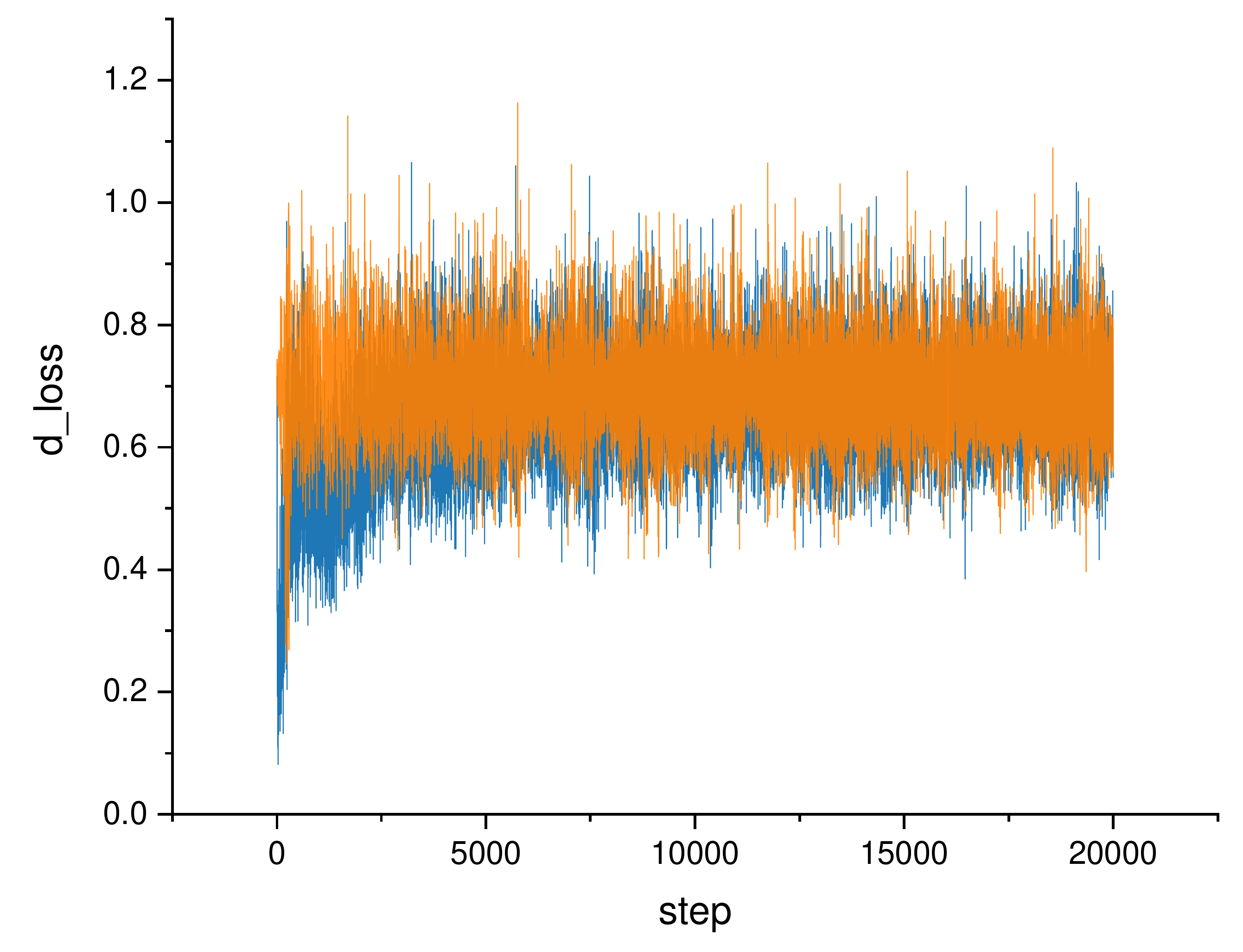} 
            & \includegraphics[width=1.25in]{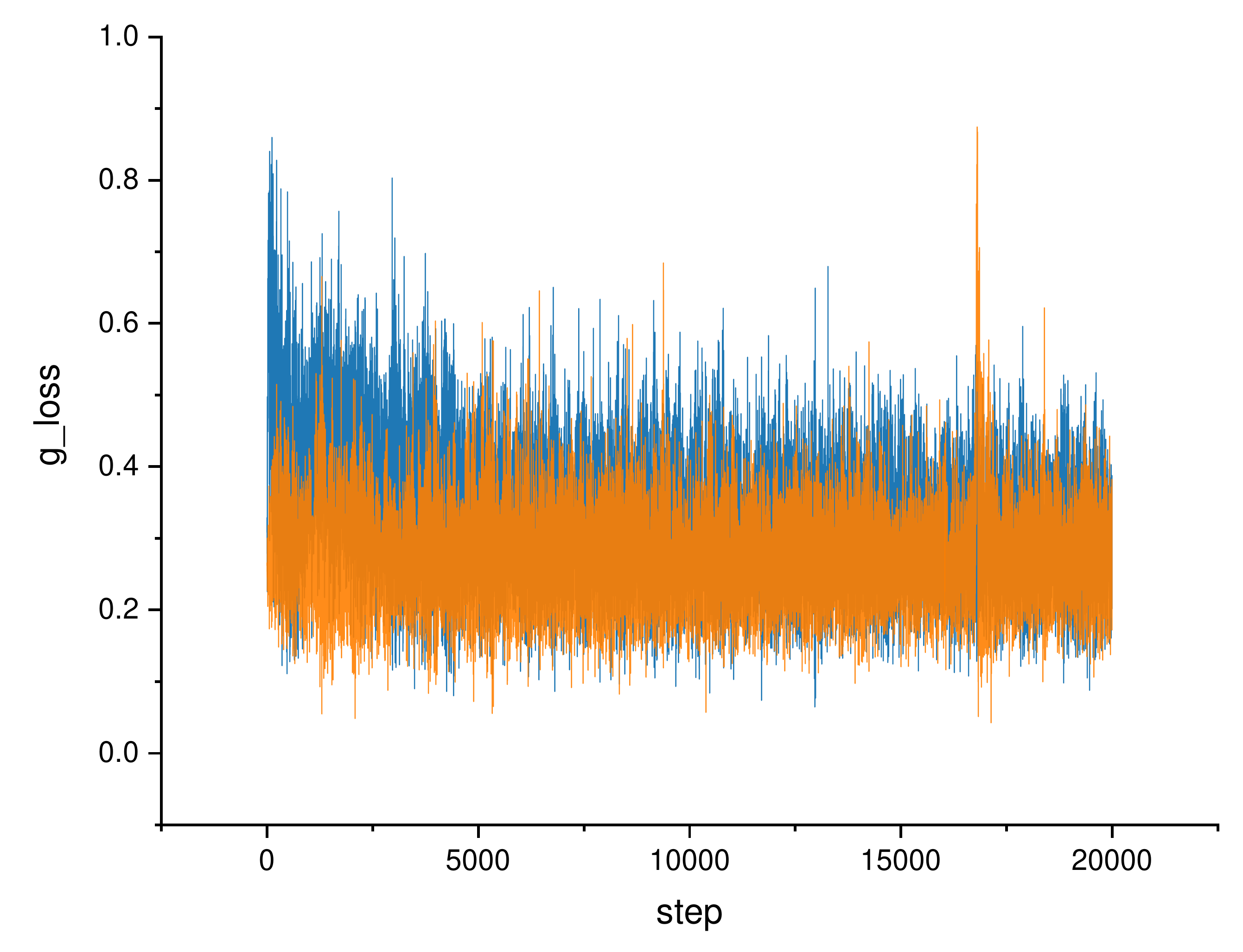} 
            & \includegraphics[width=1.25in]{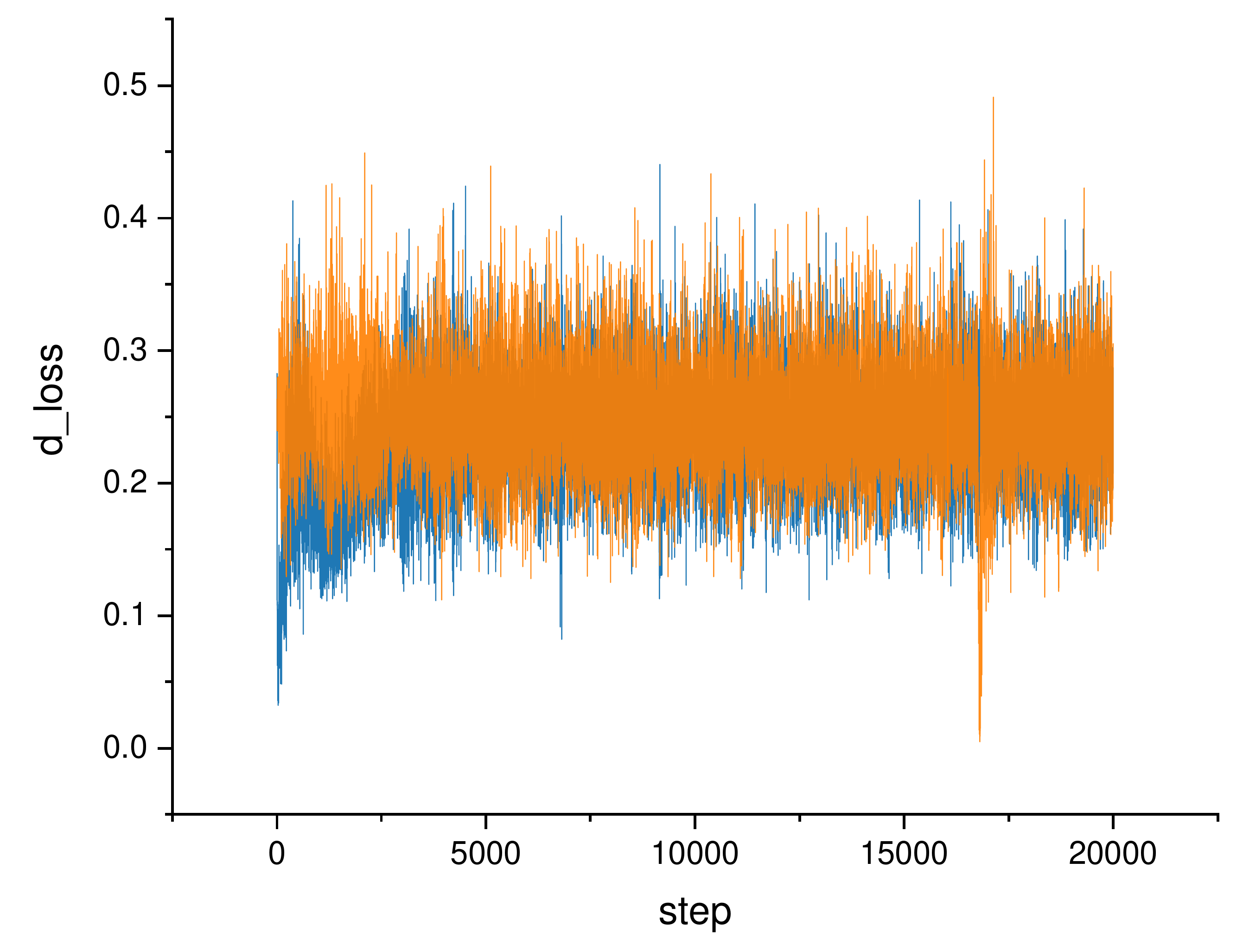} \\
            \toprule
            GAN
            & 0.8843 / 0.2668 
            & 0.6518 / 0.0965 
            & 0.3254 / 0.0928 
            & 0.2330 / 0.0421 \\
            \midrule
            Unbalanced GAN
            & 0.7370 / \textbf{0.1312} 
            & 0.6904 / \textbf{0.0755} 
            & 0.2729 / \textbf{0.0675} 
            & 0.2483 / \textbf{0.0395} \\
            \bottomrule
        \end{tabular}
    \end{center}
    \end{small}
    \vskip -0.1in
    \captionof{figure}{Loss graphs of ordinary GANs and Unbalanced GANs on the MNIST dataset. The blue line denotes the loss graph of ordinary GANs and the orange line denotes the loss graph of Unbalanced GANs. The numbers below are averages and standard deviations of each loss graph. The standard deviations of Unbalanced GANs are lower than the standard deviations of ordinary GANs.}
    \label{fig:figure2}
    \vskip 0.2in
    
    \begin{small}
    \begin{center}
        \begin{tabular}{m{0.6in}m{1.3in}m{1.3in}m{1.3in}m{1.3in}}
            \toprule
            & \multicolumn{1}{c}{Epoch 1} & 
            \multicolumn{1}{c}{Epoch 5} &
            \multicolumn{1}{c}{Epoch 10} &
            \multicolumn{1}{c}{Epoch 20}\\
            \toprule
            \multicolumn{1}{c}{WGAN} & 
            \includegraphics[width=1.3in]{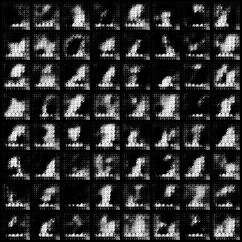} &
            \includegraphics[width=1.3in]{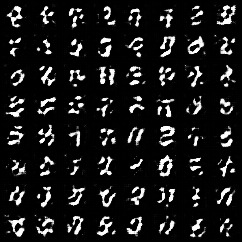} &
            \includegraphics[width=1.3in]{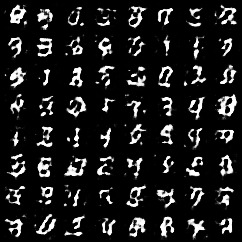} &
            \includegraphics[width=1.3in]{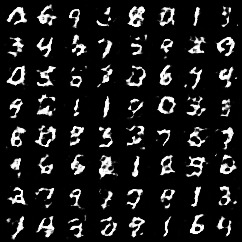} \\
            \midrule
            \multicolumn{1}{c}{\parbox{0.6in}{\centering Unbalanced\\WGAN}} &
            \includegraphics[width=1.3in]{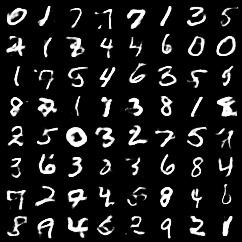} &
            \includegraphics[width=1.3in]{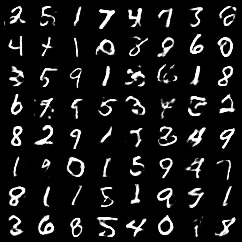} &
            \includegraphics[width=1.3in]{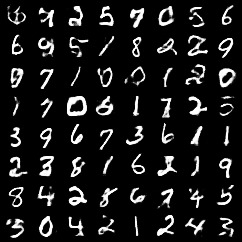} &
            \includegraphics[width=1.3in]{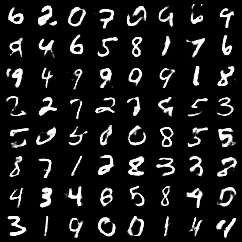} \\
            \bottomrule
        \end{tabular}
    \end{center}
    \end{small}
    \vskip -0.1in
    \captionof{figure}{Sample images generated on WGAN and Unbalanced WGAN using the MNIST dataset. Unbalanced WGAN produces better quality images at early epochs.}
    \label{fig:figure3}
\end{table*}

\subsection{MNIST}
We used 28 $\times$ 28 image size for the MNIST dataset. No data augmentation was applied to the images. Furthermore, we used 7 $\times$ 7 downsampled images for the discriminator. We made use of labels of the dataset to train in a supervised way in order to achieve better performance. \\
We assessed our models in both qualitative and quantitative ways. For quantitative analysis, we used the convergence and the standard deviation of losses since reducing the standard deviation of loss is one of the key aspects to design high-performance GANs \cite{goodfellow2016nips}. \\
We plotted the result of the experiments in Figure~\ref{fig:figure2}. As we displayed on the loss graphs, Unbalanced GANs converges faster compared to ordinary GANs. The losses of ordinary GANs oscillate at the beginning of the training and become smaller as training goes by. But the losses of Unbalanced GANs show small oscillations from the start. Additionally, the losses at the early epochs and the subsequent epochs have smaller differences and remain almost constant. Unbalanced GANs show smaller standard deviations compared to ordinary GANs. \\
For qualitative analysis, we generated sample images of WGAN and Unbalanced WGAN at every epoch. We displayed some results in Figure~\ref{fig:figure3}. At epoch 1, WGAN generated images that can be distinguished only the location of the numbers and the boundaries of the background. On the other hand, Unbalanced WGAN generated images that can be identified as a number. For other epochs, WGAN generated numerically-looking but blurry and noisy images but Unbalanced WGAN generated vivid and clear numbers.

\begin{figure*}[ht]
    \vskip 0.1in
    \begin{center}
        \subfigure[DCGAN / Unbalanced DCGAN]{\includegraphics[width=2.2in]{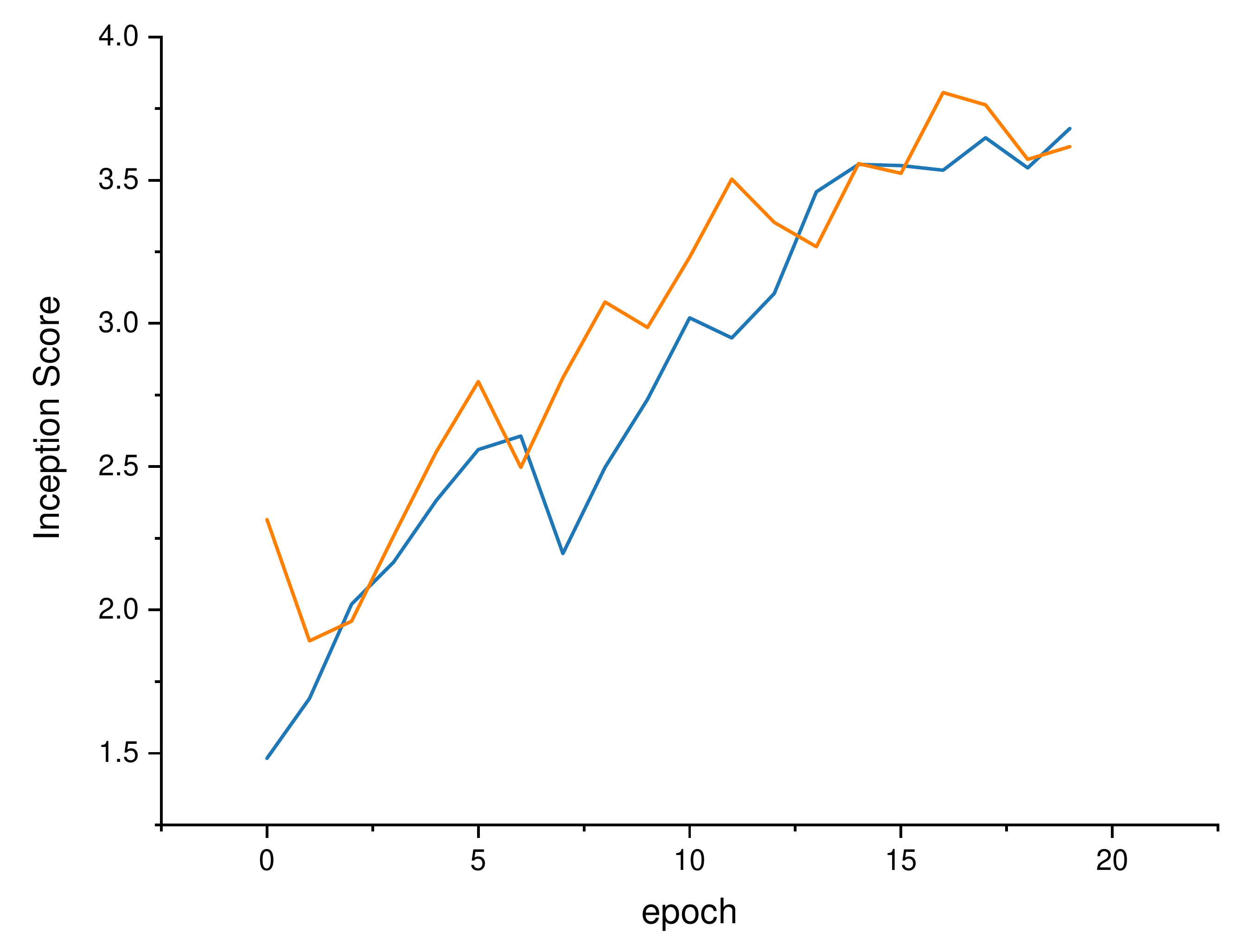}}
        \subfigure[LSGAN / Unbalanced LSGAN]{\includegraphics[width=2.2in]{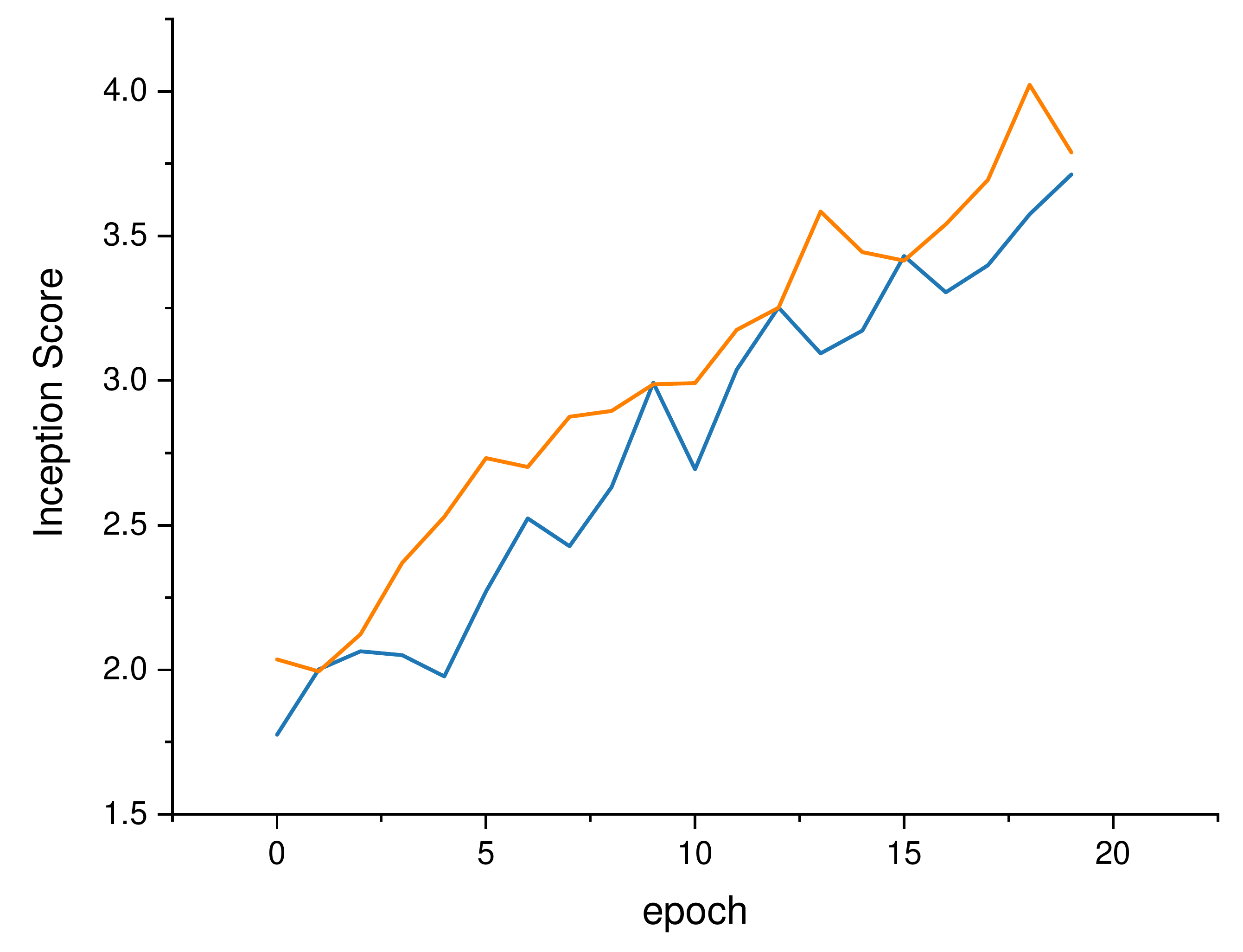}}
        \subfigure[WGAN / Unbalanced WGAN]{\includegraphics[width=2.2in]{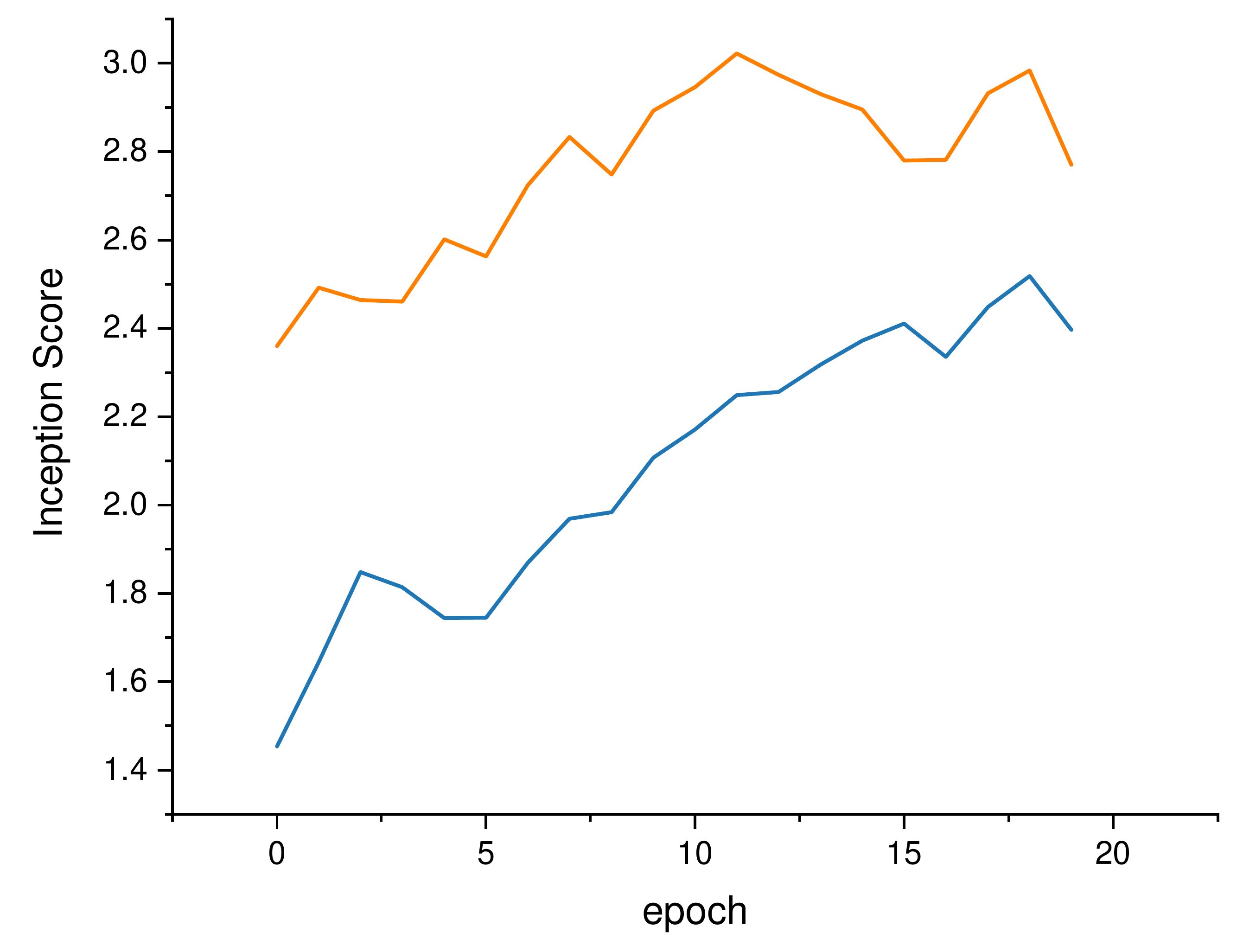}}
    \end{center}
    \vskip -0.1in
    \caption{Inception Score at each epoch on the CIFAR-10 dataset. The blue line denotes the Inception Scores of ordinary GANs and the orange line denotes the Inception Scores of Unbalanced GANs. For all networks and most of epochs, the Inception Scores of Unbalanced GANs are higher than the Inceptions Scores of ordinary GANs.}
    \label{fig:figure4}
    \vskip -0.1in
\end{figure*}

\begin{table*}[t]
    \vskip 0.1in
    \setlength\heavyrulewidth{0.25ex}
    \begin{small}
    \begin{center}
        \begin{tabular}{m{0.6in}m{1.3in}m{1.3in}m{1.3in}m{1.3in}}
            \toprule
            \multicolumn{1}{c}{Step}
            & \multicolumn{1}{c}{1000} 
            & \multicolumn{1}{c}{5000} 
            & \multicolumn{1}{c}{10000} 
            & \multicolumn{1}{c}{20000}\\
            \toprule
            \multicolumn{1}{c}{DCGAN} & 
            \includegraphics[width=1.2in]{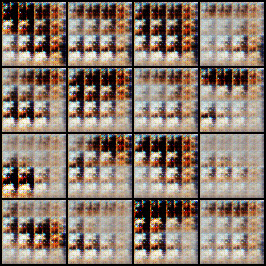} &
            \includegraphics[width=1.2in]{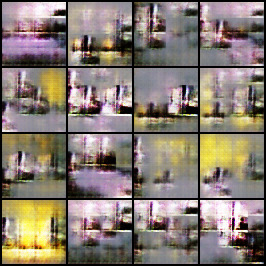} &
            \includegraphics[width=1.2in]{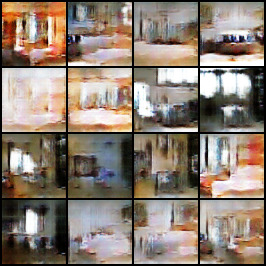} &
            \includegraphics[width=1.2in]{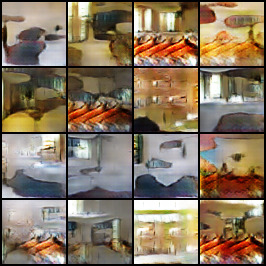} \\
            \midrule
            \multicolumn{1}{c}{\parbox{0.6in}{\centering Unbalanced\\DCGAN}} &
            \includegraphics[width=1.2in]{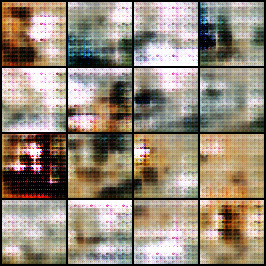} &
            \includegraphics[width=1.2in]{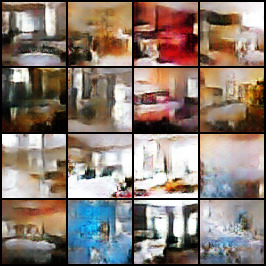} &
            \includegraphics[width=1.2in]{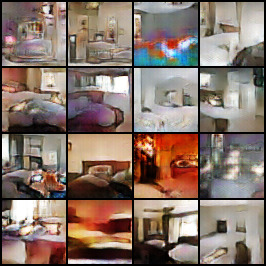} &
            \includegraphics[width=1.2in]{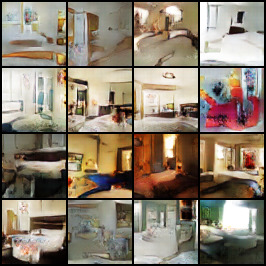} \\
            \toprule
            \multicolumn{1}{c}{LSGAN} & 
            \includegraphics[width=1.2in]{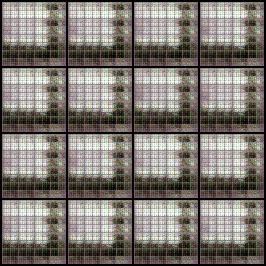} &
            \includegraphics[width=1.2in]{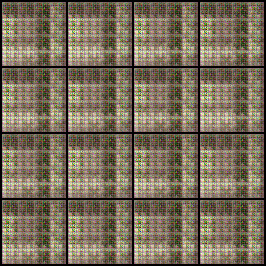} &
            \includegraphics[width=1.2in]{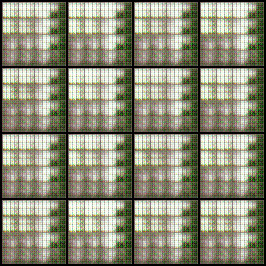} &
            \includegraphics[width=1.2in]{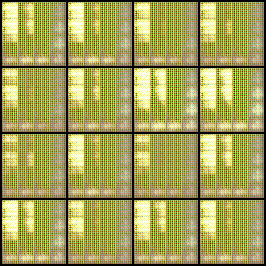} \\
            \midrule
            \multicolumn{1}{c}{\parbox{0.6in}{\centering Unbalanced\\LSGAN}} &
            \includegraphics[width=1.2in]{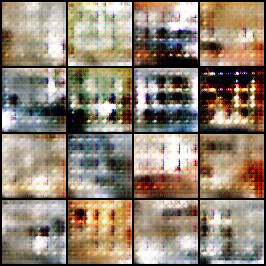} &
            \includegraphics[width=1.2in]{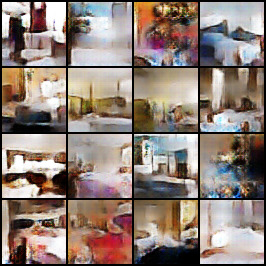} &
            \includegraphics[width=1.2in]{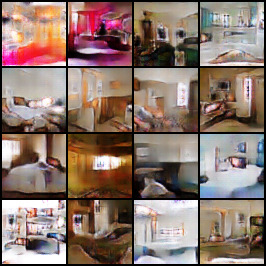} &
            \includegraphics[width=1.2in]{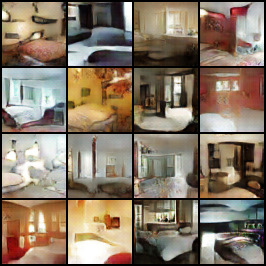} \\
            \bottomrule
        \end{tabular}
    \end{center}
    \end{small}
    \captionof{figure}{Sample images generated on ordinary GANs and Unbalanced GANs using the LSUN Bedroom dataset. Mode collapse happens in LSGAN but not in Unbalanced LSGAN. Additionally, the generated images on Unbalanced GANs are more realistic than the generated images on ordinary GANs.}
    \label{fig:figure5}
    \vskip -0.1in
\end{table*}

\subsection{CIFAR-10}
For the CIFAR-10 dataset, we used 32$\times$ 32 image size and 4 $\times$ 4 downsampled images for the discriminator. Other conditions were the same as the experiment on the MNIST dataset. According to the transferring GANs, pre-training the generator can be harmful to increase the performance of GANs. However, our experiment showed a different outcome from that. \\
We did a numerical analysis using the difference of the Inception Score \cite{salimans2016improved} of ordinary GANs and Unbalanced GANs on the CIFAR-10 dataset. The Inception Score is a metric for GANs, measuring single sample quality and diversity using the Inception model \cite{szegedy2016rethinking}. We displayed the plot of the Inception Score at each epoch in Figure~\ref{fig:figure4}. \\
Surprisingly, Unbalanced GANs get higher Inception Scores on almost all epochs. In the case of DCGAN and Unbalanced DCGAN, there are some turnovers between them but the Inception Scores of Unbalanced DCGAN are above the Inception Scores of DCGAN in broad outlines. As training goes by the Inception Scores of them become saturated and the gap between them narrows. \\
In the case of LSGAN and WGAN, the Inception Scores of Unbalanced GANs are over the Inception Scores of ordinary GANs. Unbalanced LSGAN shows small enhancement compared to LSGAN. However, Unbalanced WGAN significantly overwhelms WGAN making a huge difference in the Inception Score. \\
We did not display the losses of our experiments on the CIFAR-10 dataset but the results were similar to the experiments on the MNIST dataset. The losses converge faster and the standard deviations are lower.

\subsection{LSUN Bedroom}
We also applied Unbalanced GANs on the LSUN Bedroom dataset. We used 64 $\times$ 64 image size and for the discriminator, we downsampled until 4 $\times$ 4 image size. We trained in an unsupervised way since the dataset has only one class. \\
Unlike the MNIST dataset and the CIFAR-10 dataset, as the image size is doubled, the loss of VAE increased and took a long time to converge. Besides, the VAE pre-trained on the MNIST dataset produced less blurry real-like images, but on the LSUN Bedroom dataset produced blurry images that only overall structure can be recognized. The image size and the diversity of the dataset affected the quality of images that the variational decoder generated. \\
Because of the low quality variational decoder and excluding the conditions, the standard deviations of losses of Unbalanced GANs and ordinary GANs have almost no differences. Furthermore, the speed of convergence made no great difference. However, the quality of the generated images at early epochs was enhanced. \\
We put generated sample images of Unbalanced GANs and ordinary GANs in Figure~\ref{fig:figure5}. Unbalanced DCGAN showed better performance than DCGAN. At step 5000, DCGAN generated meaningless patterns that are not related to the bedroom. On the contrary, Unbalanced DCGAN generated blurry images that have a similar structure with the bedroom. This trend persists in subsequent steps. At step 20000, DCGAN generated images that are similar in color but distorted while Unbalanced DCGAN generated images that have similar color, structure and feature with the bedroom. \\
Especially in LSGAN and Unbalanced LSGAN, the difference grows bigger. LSGAN suffered from mode collapse at the beginning of the training and generates strange patterns. However, Unbalanced generated bedroom-like images while not suffering from mode collapse. This phenomenon shows that Unbalanced LSGAN can learn in a more stable manner than LSGAN.

\section{Conclusion}
\label{conclusion}
We introduced Unbalanced GANs which pre-trains the generator using variational autoencoder. By pre-training the generator, we can prevent the fast convergence of discriminator at early epochs and thus can balance the generator and the discriminator. Unbalanced GANs produce better performance than ordinary GANs with respect to faster convergence, low variance and better image quality at early epochs. Furthermore, we can say that the training of Unbalanced GANs is more stable than that of ordinary GANs since mode collapse happens in ordinary GANs but not in Unbalanced GANs. We believe that Unbalanced GANs can be widely applicable to other GANs to enhance their performance. It does not require a complex training process but just sequentially training VAE and GAN.\\
Our approach enables stabilized transfer learning in GANs. Using VAE rather than GAN as a pre-trained model eliminates concerns about the failure of pre-training since training VAE is not a matter of failure but time. Like Inception models, if we construct VAE that was pre-trained enough times to generate real-like images, we can utilize them to transfer knowledge to the generator and thus transfer learning can be frequently used in GANs.


\bibliography{UnbalancedGANs}
\bibliographystyle{icml2020}

\end{document}